\documentclass{article}

\usepackage[preprint]{corl_2026} % Uncomment for pre-prints (e.g., arxiv); This is like ``final'', but will remove the CORL footnote.
\usepackage{bm}
\usepackage{amsmath,amssymb}

\usepackage{booktabs}
\usepackage{graphicx}
\title{FlowMo-WM: A World Model with Object Momentum and Hidden Ambient Drift}

\author{
  Yitao Jiang\\
  Department of Computer Science \\
  Dartmouth College, Hanover, NH \\
  \texttt{yitao.jiang.gr@dartmouth.edu} \\
  \And
  Luyang Zhao \\
  Department of Electrical and Computer Engineering \\
  Clemson University, Clemson, SC \\
  \texttt{luyangz@clemson.edu} \\
  \AND
  Muhao Chen \\
  Department of Mechanical and Aerospace Engineering \\
  University of Houston, Houston, TX \\
  \texttt{muhaochen@uh.edu} \\
  \And
  Devin Balkcom \\
  Department of Computer Science \\
  Dartmouth College, Hanover, NH \\
  \texttt{devin.balkcom@dartmouth.edu} \\
}

\begin{document}
\maketitle

%===============================================================================

\begin{abstract}
World models in robot learning predict future states from visual observations and actions, enabling agents to reason about the consequences of their controls. However, many action-conditioned models are evaluated in settings where motion is dominated by immediate control, whereas aquatic surface vehicles and other real-world objects continue moving under inertia and are displaced by hidden ambient drift, such as water currents or wind. We propose FlowMo-WM, an end-to-end trainable visual world model that infers object-centric motion state and a predictive long-history context associated with hidden drift from image-action histories without direct supervision of flow fields. FlowMo-WM factorizes image-action history into a short-history latent state, trained to summarize object-centric motion, and a longer-history context, trained to summarize slowly varying exogenous influences. A zero-context residual transition separates action-conditioned base dynamics from context-dependent drift effects during latent rollout. In simulated aquatic surface-vehicle environments with diverse hidden flows, disturbances, and randomized vehicle dynamics, FlowMo-WM improves long-horizon rollout accuracy over representative action-conditioned latent world models. Prediction-time context ablations, in which the inferred context is zeroed or shuffled during rollout, show that the ambient context is important for stable prediction under hidden drift, while frozen linear probes characterize information encoded in the learned factors.
\end{abstract}


%===============================================================================

\section{Introduction}
\label{sec:introduction}

World models let robot agents predict action consequences from observations and use those predictions for control or planning \cite{ha2018worldmodels,hafner2019planet}. 
Dreamer-style latent imagination and TD-MPC-style latent dynamics have made this approach effective across many domains \cite{hafner2020dreamer,hafner2023dreamerv3,hansen2024tdmpc2}. 
DayDreamer further showed that visual world models can be trained directly on physical robots \cite{wu2023daydreamer}. 
For planning, a world model acts as a learned simulator: candidate future actions can be rolled forward, scored by predicted outcomes, and optimized before executing the first action. 
Compared with a hand-designed controller, a learned visual world model can reuse the same prediction mechanism across tasks and can plan without requiring explicit dynamics, state estimates, or disturbance measurements.

However, an important difficulty in real robotic systems is that future motion is often not determined by the current action alone. 
For aquatic surface vehicles and floating objects, motion depends on thrust, velocity, actuator lag, and ambient currents \cite{fossen2017ilos,xu2023massreview}. 
Aerial robots face an analogous problem under wind disturbances \cite{wang2024quadsurvey}. 
From images alone, a single frame reveals pose but not the full state governing future evolution, so prediction is partially observed \cite{karl2017dvbf,han2020vrm}.

Current aquatic robots already face drift as a practical control difficulty. 
SoftRafts studies floating modular robots whose motion is affected by surface drift \cite{zhao2026softrafts}. 
Flexible aquatic platforms such as the tensegrity dolphin couple body motion with the surrounding water \cite{zhao2025dolphin}.
SeePerSea shows that aquatic surface vehicles operate in visually and dynamically variable field conditions \cite{jeong2025seepersea}. 
In these systems, water can move the robot even when the commanded thrust is unchanged.

Some works show that recent trajectories can reveal hidden dynamics or task variables \cite{yu2017uposi,rakelly2019pearl}. 
Other work infers latent context in meta-RL or visual world models \cite{zintgraf2020varibad,wu2023contextwm}. 
Factorized representations can also separate controllable dynamics from exogenous variation \cite{wang2022denoisedmdp,islam2023acro}. 
Yet comparatively little visual world-modeling work has focused on objects that continue moving from their own momentum while also being displaced by hidden drift.

We propose \textbf{FlowMo-WM}, a visual world model for this regime. 
The model is trained end-to-end: the visual encoder, temporal encoders, latent transition, and pose prediction head are optimized jointly from image-action windows and future-pose supervision. 
It predicts future pose by factorizing latent dynamics into a short-history state for object-centric motion and a longer-history context for slowly varying ambient drift. 
A zero-context residual transition separates action-conditioned base dynamics from context-dependent drift effects during latent rollout.

Our contributions are: (i) a visual world-modeling formulation where object momentum and environmental transport are inferred from history; (ii) a short/long temporal factorization with a zero-context residual transition; and (iii) prediction, planning, prediction-time context-ablation, and probe evaluations in flow-affected aquatic surface-vehicle tasks.

%===============================================================================

\section{Related Work}
\label{sec:related_work}

\paragraph{Visual world models and action-conditioned prediction.}
World models support planning and policy learning through imagined rollouts from observations and actions \cite{ha2018worldmodels,hafner2019planet}. 
PlaNet and Dreamer-style models learn recurrent latent dynamics from pixels for planning or latent imagination \cite{hafner2020dreamer,hafner2023dreamerv3}. 
TD-MPC2 emphasizes compact task-oriented latent dynamics \cite{hansen2024tdmpc2}, and DayDreamer demonstrates online visual world-model learning on physical robots \cite{wu2023daydreamer}. 
Action-conditioned visual prediction has also been studied directly in pixel space for robotic manipulation \cite{finn2016video,ebert2017svp}. 
Predictive embedding methods such as I-JEPA motivate latent-space prediction without reconstructing full images \cite{assran2023ijepa}. 
LeWorldModel shows that a visual encoder and latent predictor can be trained end-to-end from pixels in a joint-embedding world-model objective \cite{maes2026leworldmodel}. 
These works provide the modeling context for our comparison models, but they do not specifically isolate the combined effect of object momentum and hidden ambient drift.

\paragraph{Partial observability, context, and factorization.}
When important state variables are not directly observed, temporal inference is required. 
Deep variational Bayes filters and recurrent latent models infer predictive hidden states from observation-action histories \cite{karl2017dvbf,han2020vrm}. 
Online system identification infers unknown physical parameters from recent trajectories \cite{yu2017uposi}. 
Meta-RL methods infer latent task variables from experience \cite{rakelly2019pearl,zintgraf2020varibad}. 
ContextWM separates context from dynamics in visual world models \cite{wu2023contextwm}. 
Related factorization work separates controllable dynamics from exogenous variation \cite{wang2022denoisedmdp,islam2023acro} or decomposes scenes through object-centric structure \cite{ferraro2025focus,feng2025fiocwm}. 
FlowMo-WM uses a simpler temporal factorization: a short window for object motion and a longer window for slowly varying drift.

\paragraph{Disturbance-aware robotics.}
Marine and aerial robots commonly face environmental disturbances from currents and wind. 
Classical marine guidance and control methods handle currents with explicit state estimates or hydrodynamic models \cite{fossen2017ilos,xu2023massreview}. 
Aerial robotics uses related disturbance-aware methods for wind \cite{wang2024quadsurvey}. 
Our learned setting is different: the model receives clean images and actions only, with no current sensor, velocity target, flow map, or privileged simulator state.
	
%===============================================================================

\section{Problem Formulation}
\label{sec:problem_formulation}

We study visual world modeling for aquatic surface vehicles whose motion is shaped by both commanded thrust and hidden environmental transport. 
In the simulator, each episode samples a boat morphology, randomized physical parameters including mass, inertia, drag, thrust scale, and actuator time constant, and one time-invariant ambient flow-field instance. 
The simulator internally evolves pose, velocity, angular velocity, actuator state, local flow, and physical parameters, but the learned world models do not receive these internal variables.

The learned models receive only clean top-down RGB images of the vehicle and the executed action history. 
The images contain no flow arrows, goal markers, velocity vectors, trajectories, or simulator overlays. 
Thus, drift and momentum can only be inferred indirectly from how the vehicle has moved under previous actions. 
FlowMo-WM is designed to use visual history to learn this predictive structure, without flow labels, velocity targets, current sensors, or privileged simulator state.

The prediction problem is to roll the future pose forward under a candidate future action sequence. 
Long-horizon accuracy is the critical quantity because small errors in hidden momentum or ambient drift can compound even when short-horizon pose prediction looks accurate.

%===============================================================================

\section{Methods}
\label{sec:method}

\begin{figure*}[t]
\centering
\includegraphics[width=\textwidth]{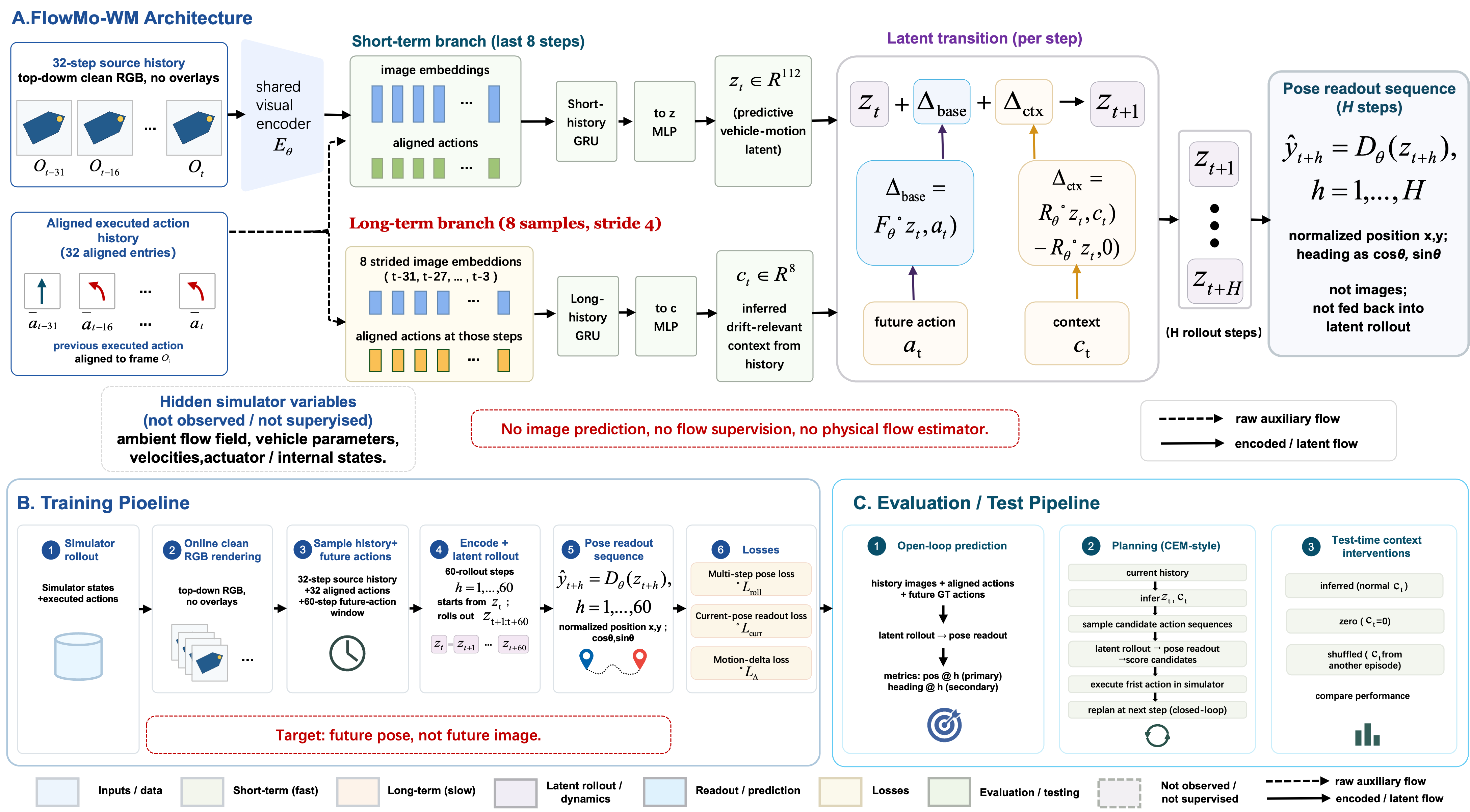}
\caption{FlowMo-WM architecture and evaluation pipeline. From a clean image-action history, a shared visual encoder feeds two temporal branches: a short-history branch infers the object-motion state $z_t$, while a longer strided branch infers the ambient context $c_t$. Future actions are rolled out in latent space with the zero-context residual transition, and a pose head predicts the normalized future pose. The lower panels summarize training with pose-rollout losses and evaluation through open-loop prediction, sampling-based planning, and prediction-time context interventions.}
\label{fig:architecture}
\end{figure*}

\subsection{FlowMo-WM Architecture}
\label{sec:flowmo_architecture}

FlowMo-WM is an end-to-end trainable visual latent world model that uses two temporal branches to organize predictive information (Fig.~\ref{fig:architecture}): a short-history object-motion state $z_t$ and a longer-history ambient-drift context $c_t$. 
Given a clean image-action history $\mathcal{H}_t=\{o_{t-31:t},a_{t-31:t}\}$ and future actions $a_{t:t+H-1}$, the model predicts normalized future poses $\hat y_{t+1:t+H}$ with $H=60$.
In implementation, the RGB image is concatenated with normalized $x$ and $y$ coordinate maps before entering the shared convolutional encoder, following CoordConv~\citep{liu2018coordconv}. 
The encoder uses four convolutional layers with SiLU activations, adaptive pooling, a linear projection, and LayerNorm. 
The short branch uses a GRU~\citep{cho2014learning} over the most recent $K_z=8$ image embeddings and actions to produce $z_t\in\mathbb{R}^{112}$, which captures pose, velocity-like motion, actuator lag, and recent control response. 
The long branch uses a separate GRU over a 32-step history with stride 4 to produce $c_t\in\mathbb{R}^{8}$, a compact summary of slowly varying drift inferred from accumulated discrepancies between commanded and observed motion. 
Flow affects training rollouts, but flow vectors or maps are never provided as inputs or targets.
The model predicts future normalized pose rather than future images, so the visual encoder is used to infer a compact dynamics state, and the readout is a pose-prediction head, not an image decoder.

Given $z_t$, $c_t$, and a candidate future action $a_t$, FlowMo-WM predicts the next latent state using
\begin{equation}
\label{eq:flowmo_transition}
  z_{t+1}
  = z_t + F_\theta(z_t,a_t)
  + R_\theta(z_t,a_t,c_t) - R_\theta(z_t,a_t,\mathbf{0}),
\end{equation}
where $F_\theta$ models action-conditioned base dynamics and $R_\theta$ models context-dependent drift. 
The subtraction makes the residual contribution exactly zero when $c_t=\mathbf{0}$, so zero-context ablations remove the learned drift contribution while preserving the base transition. 
A pose head maps latent states to normalized pose $\hat{y}_{t+h}=(\hat{x},\hat{y},\widehat{\cos\theta},\widehat{\sin\theta})$, and the transition is recursively rolled out for $H=60$ steps. 
Similar to end-to-end visual latent prediction models such as LeWorldModel~\citep{maes2026leworldmodel}, the visual encoder, latent transition, and prediction head are optimized jointly from image-action histories; unlike image reconstruction models, our supervised target is future pose. 
Training minimizes a 60-step pose rollout loss, $\sum_h \lambda_p\|\hat p_{t+h}-p_{t+h}\|_2^2+\lambda_\theta\|\hat r_{t+h}-r_{t+h}\|_2^2$, with $\lambda_p=1$ and $\lambda_\theta=2$, plus current-pose reconstruction and motion-delta losses with weights $1$ and $0.5$.

\subsection{Environment, Vehicles, and Tasks}
\label{sec:environment_vehicle_task}

To study hidden drift while keeping perception controlled, we use a two-dimensional aquatic surface-vehicle simulator in a bounded $[0,10]\times[0,10]$ workspace. 
The dynamics integrate body-frame thrust, actuator lag, anisotropic linear and quadratic drag, angular drag, inertia, and ambient flow. 
Drag is applied relative to the local flow velocity, so an unobserved current changes both passive drift and the effective response to thrust.
Simulator states are advanced with a fixed-step fourth-order Runge--Kutta integrator rather than explicit Euler integration~\citep{butcher2016numerical}, which reduces discretization error for smooth vehicle-flow dynamics but can be less advantageous when the effective vector field is non-smooth or changes abruptly within a step. 

\begin{figure*}[t]
\centering
\includegraphics[width=\textwidth]{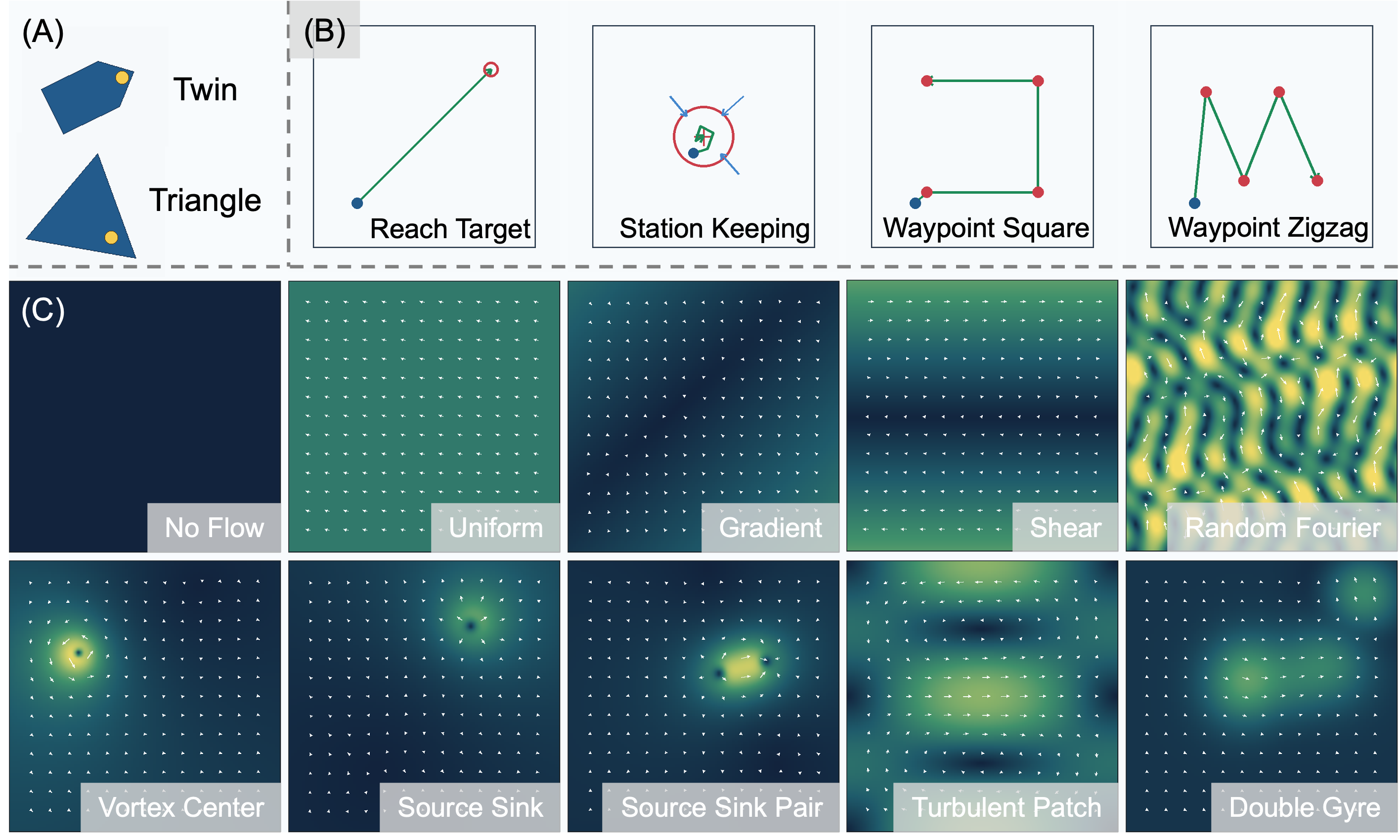}
\caption{Experimental setup and visual inputs. The figure summarizes the two vehicle morphologies, the four downstream planning tasks, and representative ambient flow families used by the simulator. Flow arrows, task paths, and goals are visualization aids for the paper; learned world models receive only clean top-down RGB images of the vehicle plus action histories. The twin-thruster boat is underactuated, while the triangular boat has three thrusters arranged to compensate for lateral drift more directly.}
\label{fig:experiment_setup}
\end{figure*}

The benchmark uses static flow fields sampled from ten families: no-flow, uniform, vortex-center, double-gyre, source-sink, source-sink-pair, gradient, shear, turbulent-patch, and random-Fourier fields. 
They cover translation, gradients, localized circulation, source/sink effects, shear, and multi-scale variation; the flow affects the simulator state but is not rendered into learning input. 
The rendered flow field in Figures~\ref{fig:experiment_setup} and~\ref{fig:qualitative_rollouts} is for visualization.

\paragraph{Vehicles and tasks.}
We use an underactuated twin-thruster boat, which controls forward motion and yaw by differential thrust, and a triangular three-thruster boat with more nearly omnidirectional actuation. 
Both sample mass, inertia, drag, thrust scale, and actuator time constant. 
For batching, actions are stored as a common length-three thruster-command vector in the 2D simulator; the twin boat uses the first two active thruster commands, while the triangular boat uses all three. 
The two morphologies probe different control structures: the twin boat must turn to redirect thrust, while the triangular boat can compensate for lateral drift more directly.
Planning tasks include reaching targets, station-keeping, square waypoints, and zigzag waypoints. 
Reach target asks the boat to drive to a fixed goal; station keeping tests whether it can hold position under drift; square and zigzag waypoints require sequential goal tracking with longer trajectories and repeated course changes.
Training and prediction evaluation were run on an NVIDIA RTX 5090 GPU.

\subsection{Training and Evaluation}
\label{sec:training_evaluation}

Training uses $2{,}400$ episodes of 300 simulator steps, corresponding to $720{,}000$ training transitions and $722{,}400$ simulator states including initial states. The held-out prediction set contains $480$ episodes, or $144{,}000$ transitions. All reported prediction and planning results are computed on held-out episodes with independently sampled initial conditions, physical parameters, action sequences, and flow-field instances; no test rollout windows are drawn from training episodes. 
All learned models receive only $160\times160$ clean top-down RGB images of the vehicle and action histories; no flow arrows, goals, velocity vectors, trajectories, or simulator overlays are rendered into the input, and RGB frames are rendered online from stored simulator state rather than stored in the dataset. 
The prediction target is pose $y_t=(x_t,y_t,\cos\theta_t,\sin\theta_t)$; positions are normalized from the $[0,10]\times[0,10]$ workspace during training, while heading is represented by cosine and sine. 
Episodes vary in vehicle morphology, physical parameters, flow field, and trajectory-generation mode, including random action, push-and-coast, passive drift, goal seeking, and waypoint following. 
With a 32-frame history and 60-step prediction horizon, the sampler yields up to $501{,}600$ valid training windows, from which the paper protocol uses $393{,}216$ windows. 
All learned models are trained for $20{,}000$ AdamW steps~\citep{loshchilov2019decoupled} with batch size $256$, learning rate $3\times10^{-4}$, and the same train/test split, renderer, window-sampling rule, and optimizer budget. 
The prediction target is pose $y_t=(x_t,y_t,\cos\theta_t,\sin\theta_t)$; positions are normalized from the $[0,10]\times[0,10]$ workspace during training, while heading is represented by cosine and sine. 
Episodes vary in vehicle morphology, physical parameters, flow field, and trajectory-generation mode, including random action, push-and-coast, passive drift, goal seeking, and waypoint following. 
All learned models are trained for $20{,}000$ AdamW steps~\citep{loshchilov2019decoupled} with batch size $256$, learning rate $3\times10^{-4}$, 32-frame history, and 60-step prediction horizon.

We compare against parameter-matched action-conditioned latent predictors adapted to our supervised pose-rollout interface: a LeWorldModel-style single-frame latent predictor~\citep{maes2026leworldmodel}, a PlaNet/RSSM-style recurrent latent model~\citep{hafner2019planet,hafner2020dreamer,hafner2023dreamerv3}, and a TD-MPC2-style compact latent dynamics model~\citep{hansen2024tdmpc2,hansen2022tdmpc}. 
These are comparison architectures rather than full agent reproductions: the LeWorldModel-style model tests current-frame latent prediction, the RSSM-style model tests generic recurrent memory, and the TD-MPC2-style model tests compact short-history latent dynamics without a separate long-context branch.
All use the same clean images, targets, optimizer budget, and no privileged flow or velocity inputs. 
Parameter counts are closely matched: FlowMo-WM has $663{,}964$ parameters, LeWorldModel-style has $664{,}612$, PlaNet/RSSM-style has $664{,}644$, and TD-MPC2-style has $667{,}780$.

Prediction evaluation uses $16{,}384$ open-loop 60-step rollout windows with ground-truth future actions. 
The primary metric is position error $\|\hat p_{t+h}-p_{t+h}\|_2$; heading error from predicted and target sine--cosine headings is computed as a secondary diagnostic, although the main figures emphasize position error and task success. 
For prediction-time context diagnostics, we also evaluate FlowMo-WM with normal context, zero context $c_t=\mathbf{0}$, and shuffled context from another sample. 
Frozen linear probes~\citep{alain2016understanding} are fit after training from $z_t$, $c_t$, and $[z_t,c_t]$ to privileged momentum and flow quantities only for analysis.

For downstream planning, learned world models score sampled candidate action sequences through latent rollout with the same route-aware planner. 
Each planning update evaluates 512 sequences over a 45-step horizon for four refinement rounds and updates the sampling distribution from the 64 best-scoring sequences.
We run four tasks, two boats, ten flow families, and 50 episodes per setting. 
Success means reaching the active goal within a radius of $0.65$; final distance is measured to the active terminal goal; the reported energy is the action-effort proxy $\sum_t\|a_t\|_2^2$ over successful episodes. 
We also include LOS references: PID/LOS~\citep{fossen2021handbook}, No-Flow LOS, Current-Estimator LOS~\citep{belleter2019observer}, and Oracle-Flow LOS. 
These are downstream references with different information assumptions, not direct learned visual world-model comparison methods.
PID/LOS uses hand-designed line-of-sight tracking from pose and goal, No-Flow LOS ignores ambient transport, Current-Estimator LOS estimates drift from recent pose history, and Oracle-Flow LOS uses privileged simulator flow as a diagnostic feed-forward signal.

%===============================================================================

\section{Results}
\label{sec:results}

\begin{figure}[t]
\centering
\includegraphics[width=\textwidth]{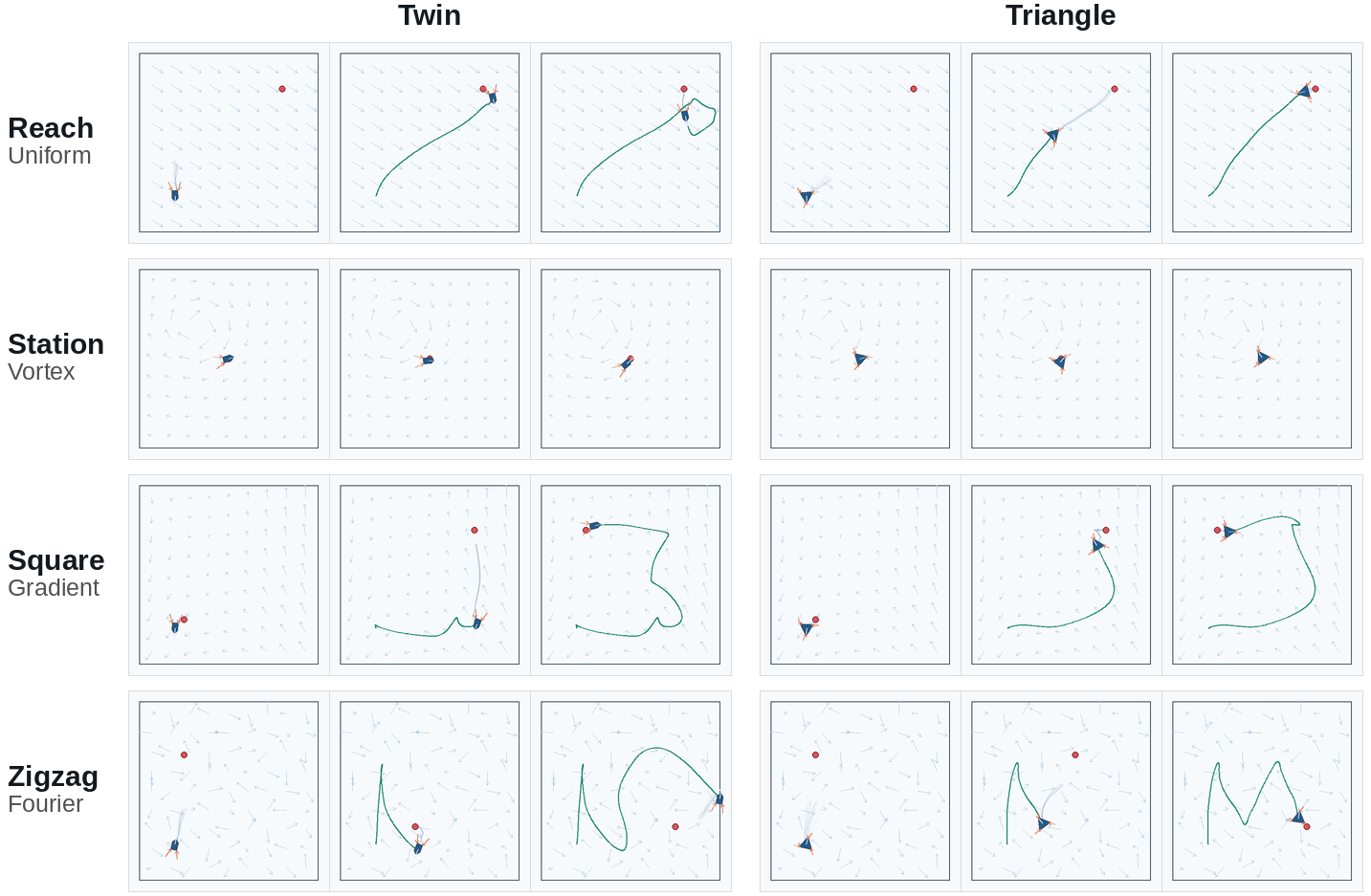}
\caption{Representative FlowMo-WM planning rollouts selected from the test executions. Rows correspond to reach, station keeping, square waypoints, and zigzag waypoints, with a different flow family selected for each task. The left group shows the twin-thruster boat, and the right group shows the triangular boat. Within each boat group, the three adjacent images are the first, middle, and final frames from one FlowMo-WM inferred-context execution. Trajectories, goals, and flow arrows are visualization overlays and are not part of the learned model input.}
\label{fig:qualitative_rollouts}
\end{figure}

\begin{figure}[t]
\centering
\includegraphics[width=\textwidth]{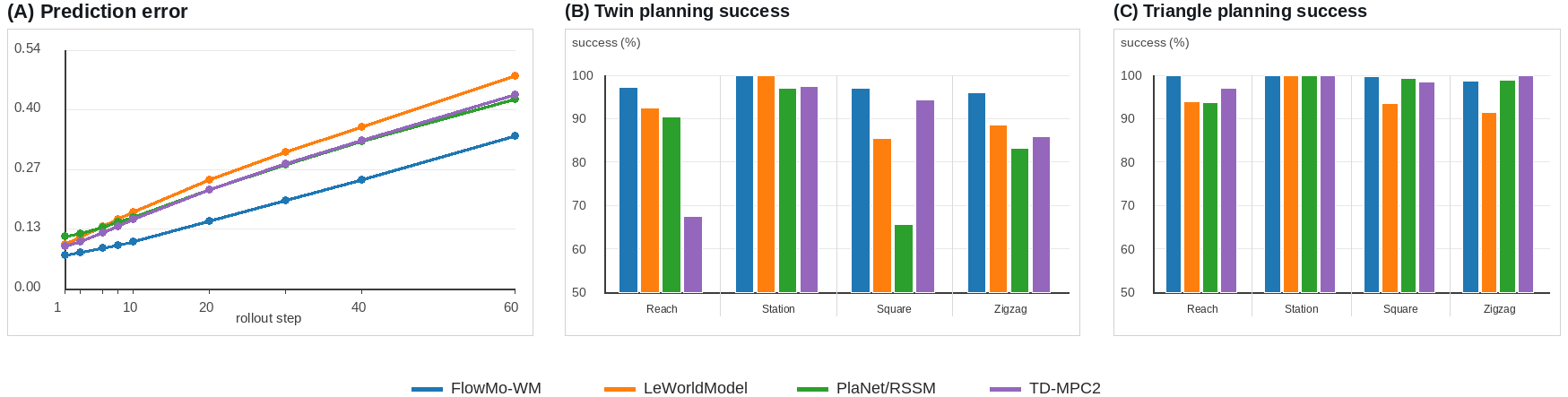}
\caption{Learned world-model prediction and planning comparison. (A) Open-loop position error over a 60-step rollout, using clean image-action history and ground-truth future actions as input; this panel averages $16{,}384$ rollout windows. (B) Downstream planning success for the twin-thruster boat, grouped by task and averaged over ten flow families; each bar aggregates $500$ episodes. (C) The same planning metric for the triangular boat, also with $500$ episodes per bar. Only learned visual world-model planners are included, so all methods use the same image inputs, planner budget, and evaluation episodes.}
\label{fig:prediction_success}
\end{figure}

Figure~\ref{fig:qualitative_rollouts} shows representative FlowMo-WM executions across tasks, boat morphologies, and flow families. 
Figure~\ref{fig:prediction_success} summarizes open-loop prediction and planning success. 
At horizon 20, FlowMo-WM reaches pos@20 $=0.151$, compared with $0.223$ for the closest comparison model TD-MPC2; at horizon 60, FlowMo-WM reaches pos@60 $=0.344$, compared with $0.426$ PlaNet/RSSM. 
This corresponds to approximately $32\%$ and $19\%$ lower position error, respectively. 
The secondary heading metric follows the same trend, with heading@60 $=0.110$ for FlowMo-WM compared with $0.122$ for the closest comparison model.

\paragraph{Prediction-time context ablations.}
With inferred context, FlowMo-WM reaches pos@20 $=0.151$ and pos@60 $=0.344$. 
Zeroing the context increases the errors to pos@20 $=0.441$ and pos@60 $=0.840$, while shuffling context from another sample further increases them to pos@20 $=0.563$ and pos@60 $=1.100$. 
Thus, removing or mismatching only the context pathway substantially destabilizes rollout.
Because the transition subtracts $R_\theta(z_t,a_t,\mathbf{0})$, the zero-context condition has a direct interpretation: it removes the learned context-dependent drift term rather than replacing the whole transition with an unrelated model.
The shuffled-context condition instead tests whether a mismatched long-history context is harmful, distinguishing useful inferred context from generic extra latent capacity.

\paragraph{Representation analysis.}
The context ablations test whether the long-history pathway is functionally used during rollout; frozen probes provide a complementary view of what information is present in the learned factors. 
Momentum is linearly decodable from $[z_t,c_t]$ with $R^2=0.667$, indicating that the learned representation contains substantial motion-state information even though velocity is never used as a training target. 
Local flow and episode-level drift have near-zero linear $R^2$, which suggests that the context improves prediction through a task-useful latent code rather than by storing a trivially linear physical current vector. 
Together, the ablations and probes support the intended role of the temporal factorization: the context branch is not just extra capacity, but a pathway whose contents affect rollout behavior.

\begin{figure}[t]
\centering
\includegraphics[width=\textwidth]{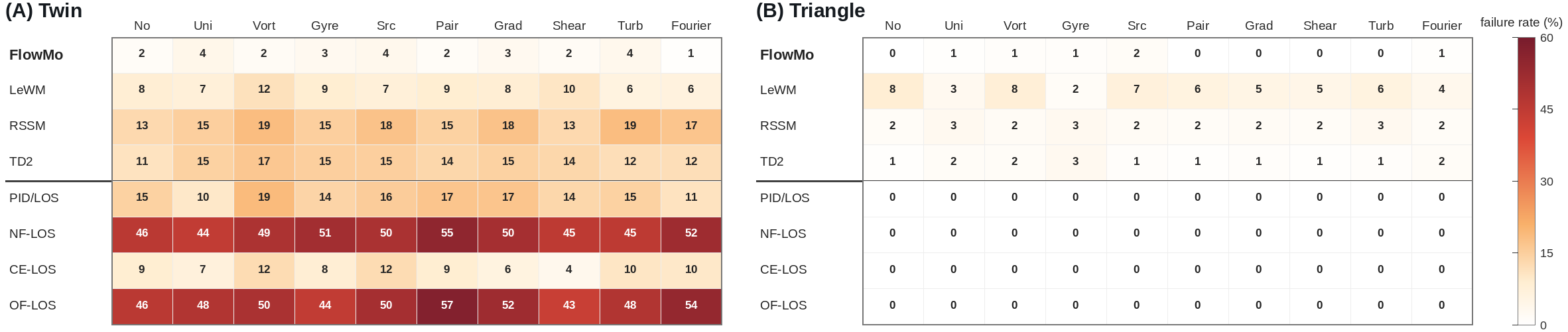}
\caption{Planning sensitivity to ambient flow family. Each cell reports failure rate, defined as $100\%$ minus success rate, aggregated across the four planning tasks for the indicated boat, method, and flow family; each cell contains $200$ episodes. Darker cells indicate more failures. The left heatmap is the underactuated twin-thruster boat, and the right heatmap is the triangular boat. Rows above the horizontal divider are learned visual world-model planners, while rows below are LOS reference controllers with different information assumptions. The failure rate is shown instead of the success rate because many settings are near saturation, making the complementary view easier to read.}
\label{fig:failure_by_flow}
\end{figure}

\paragraph{Planning results.}
Table~\ref{tab:planning_summary} reports aggregate planning results split by boat morphology. 
Within learned world-model planners, FlowMo-WM has the highest success rate, succeeding in $3940/4000$ episodes ($98.5\%$; twin: $1950/2000$, triangle: $1990/2000$), compared with $93.1\%$, $92.5\%$, and $91.0\%$ for the other learned comparison models. 
The LOS controllers use different modeling assumptions and are shown only as downstream references.
Figure~\ref{fig:failure_by_flow} breaks these planning results down by flow family. 
We plot failure rate rather than success rate because many settings are near saturation; the complementary failure view makes small but systematic flow sensitivity visible without changing the underlying metric.
The twin boat is consistently more sensitive to flow-family changes than the more directly actuated triangular boat, matching the control difficulty of compensating lateral drift with underactuated dynamics.
The LOS rows in the heatmap are not direct learned-model comparisons; they show how hand-designed references behave under their own information assumptions.

\begin{table}[t]
\centering
\caption{Aggregate downstream planning results split by boat morphology. Each method is evaluated on 4 tasks, 10 flow families, and 50 episodes per task--flow setting, giving 2{,}000 episodes per boat and method. Success is the fraction of episodes that reach the active goal. Final distance is averaged over all episodes, while energy and steps are averaged over successful episodes. Bold values mark the best entry within each block and boat column; the upper block contains learned visual world-model planners, and the lower block contains LOS reference controllers with different information assumptions.}
\label{tab:planning_summary}
\scriptsize
\setlength{\tabcolsep}{3pt}
\begin{tabular*}{\textwidth}{@{\extracolsep{\fill}}lrrrrrrrr}
\toprule
& \multicolumn{4}{c}{Twin} & \multicolumn{4}{c}{Triangle} \\
\cmidrule(lr){2-5}\cmidrule(lr){6-9}
Method & Succ. $\uparrow$ & Dist. $\downarrow$ & Energy $\downarrow$ & Steps $\downarrow$ & Succ. $\uparrow$ & Dist. $\downarrow$ & Energy $\downarrow$ & Steps $\downarrow$ \\
\midrule
FlowMo-WM & \textbf{97.5} & \textbf{0.608} & \textbf{232.9} & \textbf{268.6} & \textbf{99.5} & \textbf{0.520} & \textbf{209.4} & 218.1 \\
LeWorldModel & 91.5 & 0.790 & 343.2 & 300.9 & 94.7 & 0.701 & 327.4 & 225.7 \\
PlaNet/RSSM & 84.0 & 1.062 & 331.1 & 299.8 & 97.9 & 0.623 & 248.2 & 223.8 \\
TD-MPC2 & 86.2 & 1.065 & 300.4 & 302.6 & 98.8 & 0.564 & 274.0 & \textbf{203.1} \\
\midrule
PID/LOS & 85.4 & 0.699 & 210.5 & 300.6 & \textbf{100.0} & \textbf{0.498} & 162.4 & 320.8 \\
No-Flow LOS & 51.1 & 1.117 & \textbf{120.0} & \textbf{194.9} & \textbf{100.0} & 0.499 & 144.3 & 289.7 \\
Current-Estimator LOS & \textbf{91.2} & \textbf{0.626} & 198.7 & 298.5 & \textbf{100.0} & 0.505 & 149.3 & \textbf{286.1} \\
Oracle-Flow LOS & 50.7 & 1.128 & 131.2 & 201.1 & \textbf{100.0} & 0.513 & \textbf{141.6} & 288.8 \\
\bottomrule
\end{tabular*}
\end{table}

%===============================================================================

\section{Discussion and Limitations}
\label{sec:discussion}

FlowMo-WM targets a specific failure mode of action-conditioned visual world models: future motion may depend on hidden momentum and ambient drift that are not observable from a single image. 
The current experiments isolate this dynamics question with clean top-down rendered observations and a simplified 2D simulator; extending the method to real aquatic surface-vehicle video will require handling viewpoint changes, clutter, segmentation uncertainty, sensor noise, and richer three-dimensional disturbances. 
The dataset size should therefore be interpreted as a fixed benchmark protocol rather than an empirically optimal scaling point: it provides broad coverage over morphologies, randomized dynamics, flow families, hidden flow instances, and trajectory modes, while keeping the comparison models trained under identical data and compute budgets.
FlowMo-WM is also not expected to succeed in all flow regimes. 
When the ambient flow magnitude exceeds the vehicle's effective control authority, planning can fail even with accurate prediction; when the flow magnitude, spatial gradient, or family is outside the training distribution, the inferred context may extrapolate poorly. 
The context is inferred from a fixed input history and held constant during open-loop rollout. 
A natural extension is to add a recurrent rollout state or online context update so that environmental information can be propagated and revised during closed-loop execution, even when only a short recent history is available. 
The context ablations and probes show that the long-history pathway is useful for prediction; future diagnostics could use richer flow fields and analysis-only probes trained with ground-truth flow to test how much physical drift information is encoded. 
Finally, planning is downstream validation through a shared sampling planner, so improved optimizers or task costs may further improve control without changing the core prediction comparison.

%===============================================================================

\section{Conclusion}
\label{sec:conclusion}

We presented FlowMo-WM, a visual latent world model for systems whose future motion is shaped by both object momentum and hidden ambient drift. 
The method factorizes image-action history into a short-history object-motion state and a longer-history ambient context, then uses a zero-context residual transition to separate action-conditioned base dynamics from context-dependent drift effects. 
Across aquatic surface-vehicle prediction and planning experiments, FlowMo-WM improves long-horizon rollout accuracy over parameter-matched action-conditioned latent world-model comparison models, and prediction-time context ablations show that the inferred ambient context is functionally important for stable prediction. 
These results suggest that visual world models for open physical environments should represent not only commanded control, but also how objects continue moving and how unobserved surroundings transport them. 

%===============================================================================

\clearpage
% The acknowledgments are automatically included only in the final and preprint versions of the paper.
\acknowledgments{Thanks Yewei Huang and Alberto Quattrini Li for providing learning workstation for this project.}

%===============================================================================

\bibliography{reference} % .bib

\end{document}